# An Analytical Study of different Document Image Binarization Methods

Mahua Nandy (Pal) and Satadal Saha*
*Department of CSE, MCKV Institute of Engineering, Howrah, India*
* Corresponding author. Email: satadalsaha@yahoo.com

*Abstract*—**Document image has been the area of research for a couple of decades because of its potential application in the area of text recognition, line recognition or any other shape recognition from the image. For most of these purposes binarization of image becomes mandatory as far as recognition is concerned. Throughout couple decades standard algorithms have already been developed for this purpose. Some of these algorithms are applicable to degraded image also. Our objective behind this work is to study the existing techniques, compare them in view of advantages and disadvantages and modify some of these algorithms to optimize time or performance.**

## I. INTRODUCTION

DOCUMENTS are normally stored in gray level format having a maximum of 256 different gray values (0 to 255). To extract some information from the document image it is required to be processed number of times. Especially, if we want to recognize the image or part of it then binary image seems to more useful than gray level image. Bi-level information embedded in a binary image decreases the computational load for processing and also it enables us to use simplified analytical steps rather than 256-level gray scale image.

Binarization is the method of converting any gray scale image (popularly known as multi-tone image) into black-and-white image (popularly known as two-tone image). This conversion is based on finding a threshold gray value and deciding whether a pixel having a particular gray value is to be converted to black or white. Usually within an image the pixels having gray value greater than the threshold is transformed to white and the pixels having gray value lesser than the threshold is transformed to black. Binarization has been the area of research for last fifteen years or so mainly to find the threshold value for any image. Most of the algorithms till developed are of generic type using statistical parameters computed over the image with or without using local information or special content within the image.

The most convenient and primitive method is to find a global threshold for the whole image and binarizing the image using the single threshold. In this technique the local variations are actually suppressed or lost, though they may have important information content. On the other hand, in case of determining the threshold locally, a window is used around a pixel and threshold value is calculated for the window. Now depending on whether the threshold is to be used for the center pixel of the window or for the whole window, the binarization is done on pixel-by-pixel basis, where each pixel may have a calculated threshold value, or on region-by-region basis where all pixels in a region or window may have a threshold value.

In most of the practical cases, the binarization method fails because of the degradation of the image. The degradation may occur due to the poor method of acquisition of image or due to poor quality of original source. Degradation may also occur due to non-uniform illumination over the original source. The major contribution of research for binarization is to recover or extract information from a degraded image. Otsu [1] developed a method based on gray level histogram and maximizes the intra-class variance to total variance. Sauvola [2] developed an algorithm for text and picture segmentation within an image and binarized the image using local threshold. Gatos [3] used Wiener filter and Sauvola's adaptive binarization method. In the work presented in [4] also Sauvola's adaptive thresholding is used for binarization. Valverde [5] binariesd the image using Niblack's technique. A slight modification of Niblack's method is done in [6] by Zhang.

In the present work, we have implemented binarization technique following Otsu, Sauvola, Bernsen and Co-occurrence matrix methods. Some of the methods use global threshold and others use local threshold. We have also studied the advantages and drawbacks of the aforesaid methods we have implemented. After that we have used sliding window concept to introduce modification in Sauvola's method to make it faster and modification is also done on Bernsen's method to use standard deviation in it and the method is improved by applying Otsu's method locally.

## II. DATASET GENERATION

Any theory can be established through experiments only and preparation of dataset plays a vital role towards the success of the experiments. In the present work, we have applied the algorithms over document images. For this purpose, pages from story book, news paper, bound volume and written notes are scanned through a scanner and stored in the disk in convenient structural format. Over 100 images have been created for the experiment. We call these images as raw images or the original images. For the exhaustive study of the algorithms we have processed some of the images to introduce error in it. For example, some of the images are blurred, some of them are sharpened and some of them are made noisy. We have introduced salt and pepper noise only for introduction of noise. To make a complete dataset, 20 images are kept in original quality, 20 images are made blurred only, 20 images are made sharpened only and 20 images are made noisy only. We have also mixed up the impairments to make the images more degraded. In the process, we have made 40 images blurred and noisy, 40 images sharpened and noisy and for 10 images we have made the image blurred, noisy and then again have made it sharpened. So with these combinations we have run the algorithms on a dataset of total 170 images.

## III. PRESENT WORK

We have focused our work mainly on the study and implementation of different methods of binarization, comparison of the methods and modification of the algorithms to get better result. Following subsections describes the methods we have implemented.

### 3.1. Binarization using Otsu algorithm

#### 3.1.1. Standard Otsu or Global Otsu

A nonparametric and unsupervised method of automatic threshold selection for picture segmentation is presented by **Nobuyuki Otsu** who proposed an algorithm for threshold selection.

Let the pixels of a given image be represented in $L$ gray levels $[1, 2, ......, L]$. The no. of pixels in gray level $i$ is denoted by $n_i$ and the total no. of pixels can be expressed as $N = n_1 + n_2 + ...... + n_L$. Then suppose that the pixels were dichotomized into two classes $C_0$ and $C_1$, which denotes pixels with levels $[1,.....,k]$ and $[k+1,......,L]$, respectively. This method is based on a discriminant criterion which is the ratio of between-class variance and total variance of gray levels. The between-class variance is given by

$$\sigma_B^2 = \omega_0 \omega_1 (\mu_1 - \mu_0)^2$$

The total-*class* variance is given by

$$\sigma_T^2 = \sum (i - \mu_T)^2 p_i$$

where, $p_i = (n_i / N)$ is the probability distribution of gray level $i$.

The discriminant criterion is given by

$$\eta = \frac{\sigma_B^2}{\sigma_T^2}$$

The probabilities of class occurrence and the class mean levels respectively are

$$\omega_0 = \Pr(C_0) = \sum p_i = \omega(k)$$

$$\omega_1 = \Pr(C_1) = \sum p_i = 1 - \omega(k)$$

The $0^{th}$ order cumulative moments of the histogram upto the $k^{th}$ level is given by

$$\mu_0 = \sum_{i=1}^{k} \frac{ip_i}{\omega_0}$$

The $1^{st}$ order cumulative moments of the histogram upto the $k^{th}$ level is given by

$$\mu_1 = \sum_{i=k+1}^{L} \frac{ip_i}{\omega_1}$$

Total mean level of the original image is given by

$$\mu_T = \sum_{i=1}^{L} ip_i$$

The optimal threshold of an image maximizes $\sigma_B^2$ which in turn maximizes the separability of the resultant classes in gray levels.

#### 3.1.2. Local Otsu

In global Otsu, a global threshold value is generated using which binarization is done over the whole image. Sometimes this suppresses the local variation within the image. To make the method more effective Otsu's method is applied locally or over small windows and for $n$ number of windows $n$ number of thresholds are generated and binarization is done in each window using local threshold. In the present work, we have created nine equidimensional windows over the image and applied the Otsu algorithm over each window.

| Window1 Threshold=th$_1$ | Window1 Threshold=th$_2$ | Window1 Threshold=th$_3$ |
|---|---|---|
| Window1 Threshold=th$_4$ | Window1 Threshold=th$_5$ | Window1 Threshold=th$_6$ |
| Window1 Threshold=th$_7$ | Window1 Threshold=th$_8$ | Window1 Threshold=th$_9$ |

Fig. 1. Window based Local Otsu generates local thresholds

### 3.2. Binarization using Sauvola algorithm

Another nonparametric and unsupervised method of automatic threshold selection for text and picture segmentation is presented by Sauvola. This method is highly efficient and can binarize the image after segmenting it into two classes: picture region and text region.

In this method, neighbors of each pixel are considered to determine whether the central pixel will become black or white. The formula used for this purpose is the following:

$$T(x,y) = m(x,y)\left[1 + k\left(\frac{s(x,y)}{R} - 1\right)\right]$$

where, $m$ is the mean gray level in the considered window, $s$ is standard deviation of the gray levels in the window, $R$ is the dynamic range of the variance and $k$ is a constant (usually taken as 0.5).

Now for each pixel, if the gray value is greater than the threshold then it is converted to white pixel, otherwise it is converted to black pixel.

### 3.3. Binarization using Bernsen algorithm

Bernsen's thresholding method computes the local maximum and minimum for each pixel around the neighborhood that pixel. Let $f(x,y)$ be the gray value of a pixel, with gray values in the range $[0, 1, ……., L-1]$. Among the neighbors the maximum ($f_{max}(x,y)$) and the minimum ($f_{min}(x,y)$) gray values are searched and the median of these two values is considered as the threshold for the pixel in consideration.

$$g(x,y) = \frac{f_{max}(x,y) + f_{min}(x,y)}{2}$$

This gray value $g(x,y)$ acts as a threshold for binarization as mentioned below.

$b(x,y) = 1$, if $f(x,y) < g(x,y)$
$= 0$, otherwise.

To avoid *Ghost Phenomena*, the local variance $c(x,y)$ can be computed as

$$c(x,y) = f_{max}(x,y) - f_{min}(x,y)$$

The classification of a foreground pixel $f(x,y)$ can be verified by examining the local variance being higher than a threshold. Each pixel is classified as object or background pixel according to the following condition:

$b(x,y) = 1$, if( $f(x,y) < g(x,y)$ and $c(x,y) > c^*$ )
    or $f(x,y) < f^*$ and $c(x,y) <= c^*$)
  $= 0$, otherwise

The thresholds $c^*$ and $f^*$ are determined by applying Otsu's method to the histogram of $c(x,y)$ and $g(x,y)$ respectively.

### 3.4. Binarization using Co-occurrence matrix algorithm

Co-occurrence matrix is extensively used in texture analysis. The method is based on the estimation of the second order joint conditional probability density function $P_{d,r}(i,j)$, which is a 2D matrix. Each $P_{d,r}(i,j)$ is the probability of going from a gray level $i$ to a gray level $j$ in a given direction $r$ at a given inter sample spacing d. The co-occurrence matrix $P_{d,r}$ is a representation of the estimated values. It is square matrix of dimension $N_g$, which is the number of gray levels in the image.

Two co-occurrence matrices are derived from the document image, for $r=0º$ and $r=90º$, and $d=3$, then averaged together. A histogram is constructed for the gray levels running from $0$ to $(N_g-1)$. The cumulative frequency of occurring at each level x is defined by the summation of all matrix elements with rounded up value of $(i+j)/2$ equal to x, where $i$ and $j$ are row and column positions of the matrix. Then Otsu algorithm is applied globally to find out the threshold value for binarization of the whole image.

### 3.5. Our approach: Faster Sauvola algorithm

The original Sauvola's method is based on sliding window principle in which for calculating the threshold for each pixel, the neighbors of it are considered for calculation. Then for calculating the same for the next pixel the current neighbors are considered afresh for calculation, though part of them are already considered for calculating the threshold for the previous pixel. In the present work, whenever the window is shifted from one position to the other column wise, we have eliminated only the first column values from the previous result and added the values of the newly added column into the result for calculating the threshold for the current pixel. Similarly, whenever the window is shifted from one row to the other row wise, we have eliminated only the first row values from the previous result and added the values of the newly added row into the result for calculating the threshold for the current pixel.

### 3.6. Our approach: Modified Bernsen algorithm

Bernsen algorithm is already discussed in section 3.3. In the method for avoiding the ghost phenomenon the local variance $c(x,y)$ is calculated simply by computing the difference between the highest and lowest gray values in a window around a center pixel. In the present work, we have calculated the $c(x,y)$ by computing the standard deviation of the gray values of the pixels in a window around a centre pixel using the following formula.

$$c(x,y) = \sqrt{\frac{\sum (f(x_i, y_i) - f_{mean}(x,y))^2}{N}}$$

where, $f(x_i,y_i)$ indicates the gray values of the $(x_i,y_i)^{th}$ pixel in the window, $f_{mean}(x,y)$ is the mean gray value in a window around the $(x,y)^{th}$ pixel and $N$ is the total number of pixel in the window.

Local Otsu is applied on the $c$ and $g$ matrices to get local thresholds for the window. Now these local thresholds are used in the respective windows of the original gray image for binarizing the pixels in that window.

IV.  EXPERIMENTAL RESULTS AND DISCUSSIONS

We have run the algorithms for the methods discussed in section 3.1. to 3.6. Fig. 2. shows some of the original images over which the methods are applied upon.

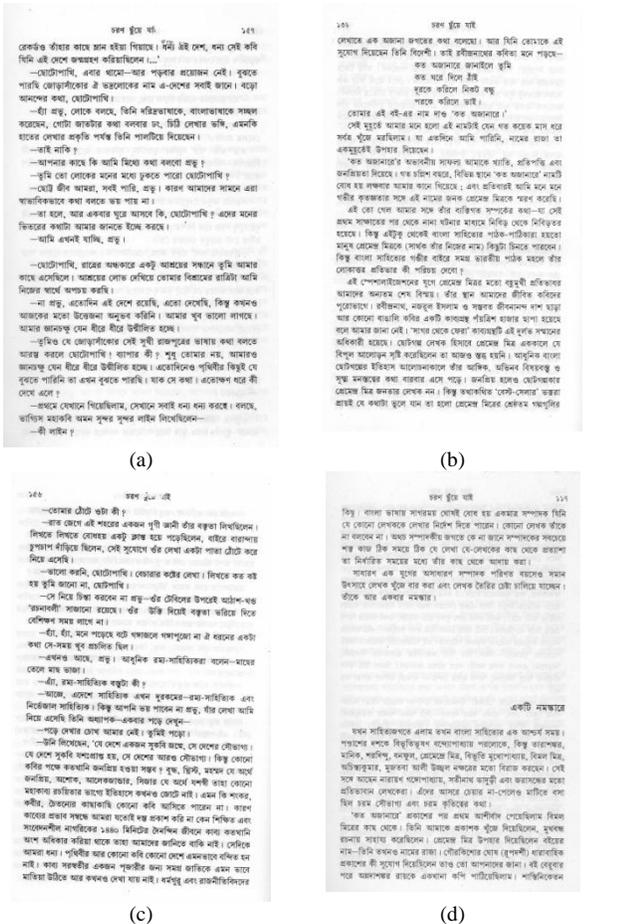

(a)    (b)

(c)    (d)

Fig. 2(a-d). Original document images

Fig.3. shows the effect of binarization through the methods of Otsu and Sauvola, when applied on a blurred image shown in Fig. 3(b).

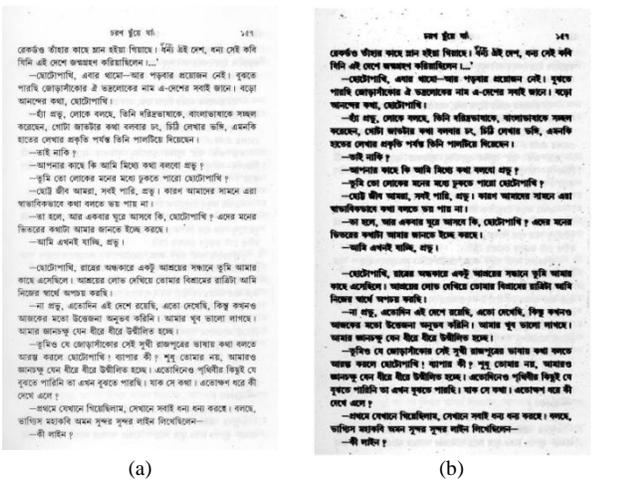

(a)    (b)

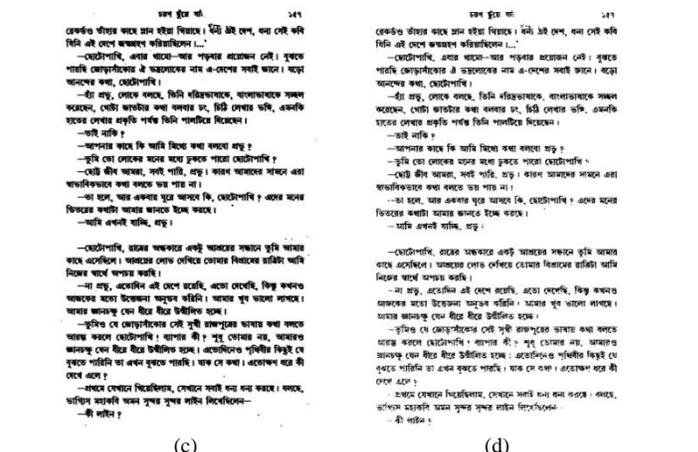

(c)    (d)

Fig. 3. (a) Original image, (b) Image in (a) after blurring, (c) Otsu output for the blurred image, (d) Faster Sauvola output for the blurred image

In Fig. 4 the binarization through different methods are shown. It is seen that Sauvola gives better result than others, where as modified Bernsen gives better result than original Bernsen. In case of Otsu also the noise could not be removed totally.

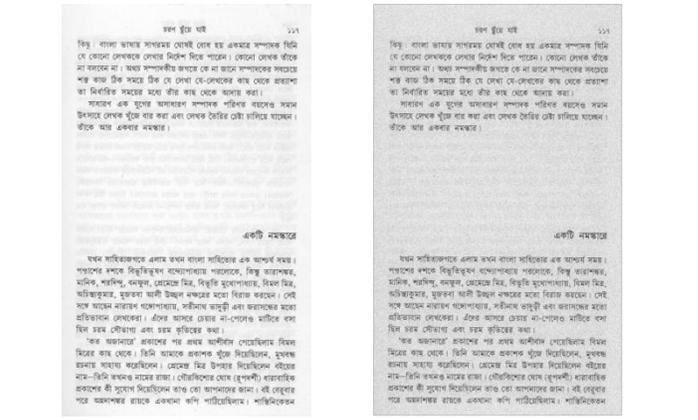

(a)    (b)

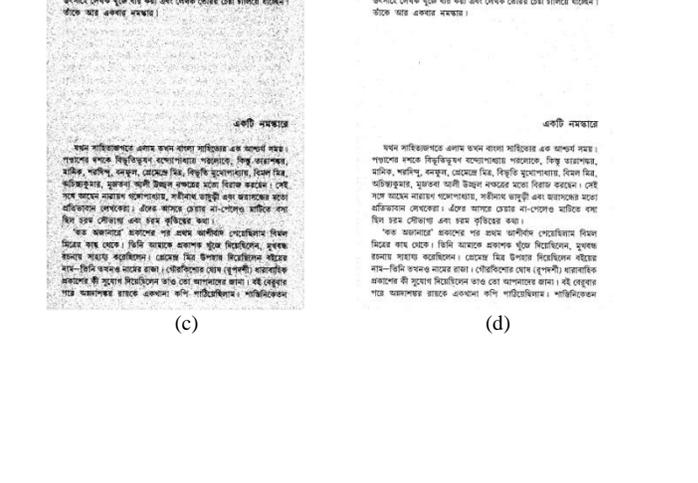

(c)    (d)

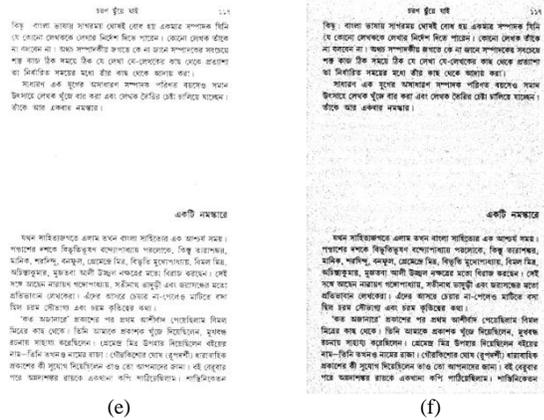

(e)           (f)

Fig. 4. (a) Original image, (b) Image in (a) after noise introduction, (c) Otsu output for the noisy image, (d) Faster Sauvola output for the noisy image, (e) Bernsen output for the noisy image, (f) Co-occurrence matrix output for the noisy image

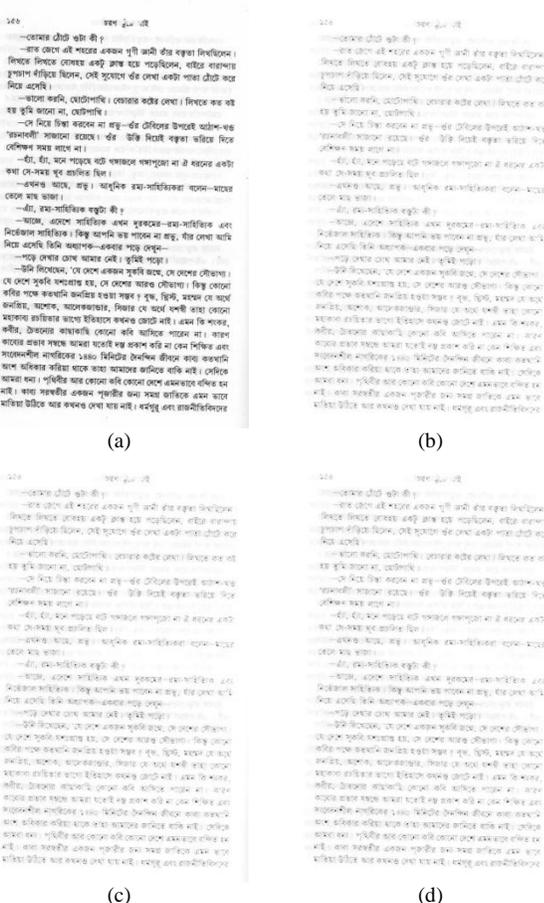

(a)           (b)

(c)           (d)

Fig. 5. (a) Original image, (b) Image in (a) after noise sharpening, (c) Otsu output for the sharpened image, (d) Faster Sauvola output for the sharpened image

When the methods are applied on a sharpened image they fail to binarize the image with high efficiency, as seen from Fig. 5.

## V. CONCLUSION

From the exhaustive study it is seen that in general Otsu is the base method for binarization if we use local threshold instead of global threshold though this technique fails to provide good binarization in case of noisy images. Sauvola's method works well in case of noisy document images but its performance degrades in case of comparatively clear images. The method itself is computationally very heavy. In our approach we have modified the original Sauvola and its performance has become many folds faster. Original Bernson's method is found to be somewhat less efficient as far as binarization is concerned as it uses global threshold. But with the use of standard deviation instead of using difference of maximum and minimum gray values and using Otsu locally in the method the algorithm is found to perform better.

## ACKNOWLEDGEMENT

We are grateful to MCKV Institute of Engineering and specifically to the Department of Computer Science & Engineering for providing necessary atmosphere to carry on the research work.

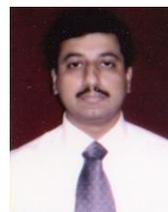

Satadal Saha was born in Kolkata, India in 1974. He has done B. Sc. (Physics Hon's) from Bidhannagar College. Kolkata in 1995, completed B. Tech. in Applied Physics and M. Tech. in Optics and Optoelectronics from University of Calcutta in 1998 and 2000 respectively.

In 2000, he joined as a Project Fellow in the Department of CST in BESU (formerly B. E. College), Howrah. In 2001, he joined as a Lecturer in the Department of Information Technology, Govt. College of Engg. and Textile Technology (formerly known as College of Textile Technology), Serampore. In 2004, he joined as a Lecturer in the Department of Computer Science and Engg., MCKV Institute of Engg, Howrah and he is continuing his service there as a Sr. Lecturer. He has published a book titled Computer Network (New Delhi: Dhanpat Rai and Co. Ltd., 2008). His research areas of interest are image processing and pattern recognition.

Mr. Saha is a member of IETE and CSI. He was also a member of the executive committee of IETE, Kolkata for the session 2006-08.